\documentclass{llncs}
\usepackage{times}
\usepackage{epsfig}
\usepackage{graphicx}
\usepackage{algorithm}
\usepackage{algpseudocode}
\usepackage{amsmath}
\usepackage{amssymb}
\usepackage{mathtools}

\usepackage{subfigure}
\usepackage{caption} 
\usepackage{enumitem}

\usepackage{color}

\usepackage{setspace}
\begin{document}
\frontmatter          
\pagestyle{headings}  
\mainmatter              
\title{Cost-Sensitive Active Learning for Intracranial Hemorrhage Detection}

\author{Weicheng Kuo\inst{1} \and
Christian H{\"a}ne\inst{1}\and
Esther Yuh \inst{2} \and
Pratik Mukherjee \inst{2}\and
Jitendra~Malik\inst{1}}
\authorrunning{Weicheng Kuo et al.} 
%
%
\institute{University of California Berkeley, Berkeley CA, 94720, USA,\\
\and
University of California San Francisco School of Medicine,\\
San Francisco CA, 94143, USA,\\
\email{wckuo@berkeley.edu}}

\maketitle              

\begin{abstract}

Deep learning for clinical applications is subject to stringent performance requirements, which raises a need for large labeled datasets. However, the enormous cost of labeling medical data makes this challenging. In this paper, we build a cost-sensitive active learning system for the problem of intracranial hemorrhage detection and segmentation on head computed tomography (CT).  We show that our ensemble method compares favorably with the state-of-the-art, while running faster and using less memory. Moreover, our experiments are done using a substantially larger dataset than earlier papers on this topic. Since the labeling time could vary tremendously across examples, we model the labeling time and optimize the return on investment. We validate this idea by core-set selection on our large labeled dataset and by growing it with data from the wild. 

\end{abstract}

\vspace*{-0.5cm}
\section{Introduction}
\vspace*{-0.2cm}

Clinical applications set very high bars for machine learning algorithms, because any misdiagnosis could impact treatment plans and gravely harm the patient. 
To reach the required performance, supervised learning is the leading technique, and its success is well established. However, a challenge in supervised learning is that it requires a large amount of labeled data, especially when deep neural networks are used. 
Unfortunately, expert labeling of medical images requires enormous time and cost. The problem is exacerbated when accurate pixelwise labeling is required. Accordingly, medical segmentation datasets tend to be relatively small~\cite{sirinukunwattana2017gland,zhang2016coarse}.  

Active learning (AL) aims to address the paucity of labeled data by reasoned choice of which available unlabeled examples to annotate~\cite{settles2008active,yang2017suggestive,seung1992query,lewis1994sequential,mahapatra2013semi}. A limitation of many prior studies of AL is that they validated AL only in a core-set selection setting,~\cite{sener2018active} 
rather than demonstrating its utility in growing the labeled data, and also did not attempt to model the cost of labeling~\cite{settles2008active,yang2017suggestive,mahapatra2013semi}. However, the potential value/use of AL is not in achieving comparable performance with less data, but in improving the model while also minimizing labeling costs.  On other problems it has been shown that labeling costs vary greatly from one example to another~\cite{settles2008active,settles2012active,tomanek2010resource}. In the case of intracranial hemorrhage, we observe that times needed for pixelwise labeling vary up to 3 orders of magnitude for different cases (See Fig.~\ref{fig:pred_label_time}). Most AL studies to date select examples without addressing this wide variation in labeling time~\cite{yang2017suggestive,seung1992query,lewis1994sequential,sener2018active,mahapatra2013semi}.

In this paper, we propose a cost-sensitive AL system by combining the query-by-committee~\cite{seung1992query} approach with labeling time prediction for each example. Our uniform-cost AL system compares favorably with the state of the art~\cite{yang2017suggestive}, while the cost-sensitive system gives a further boost under labeling time constraints. All experiments are conducted on our pixelwise-labeled dataset ($29095$ frames), which is about two orders of magnitude larger than standard MICCAI segmentation datasets~\cite{sirinukunwattana2017gland,zhang2016coarse}.
Moreover, our system is simpler, faster, and uses less memory than earlier works~\cite{yang2017suggestive,sener2018active}. Through the example of intracranial hemorrhage detection, we demonstrate the potential of cost-sensitive active learning to scale up medical datasets efficiently.  

\begin{figure}[t]
    \centering
    \includegraphics[width=0.9\linewidth]{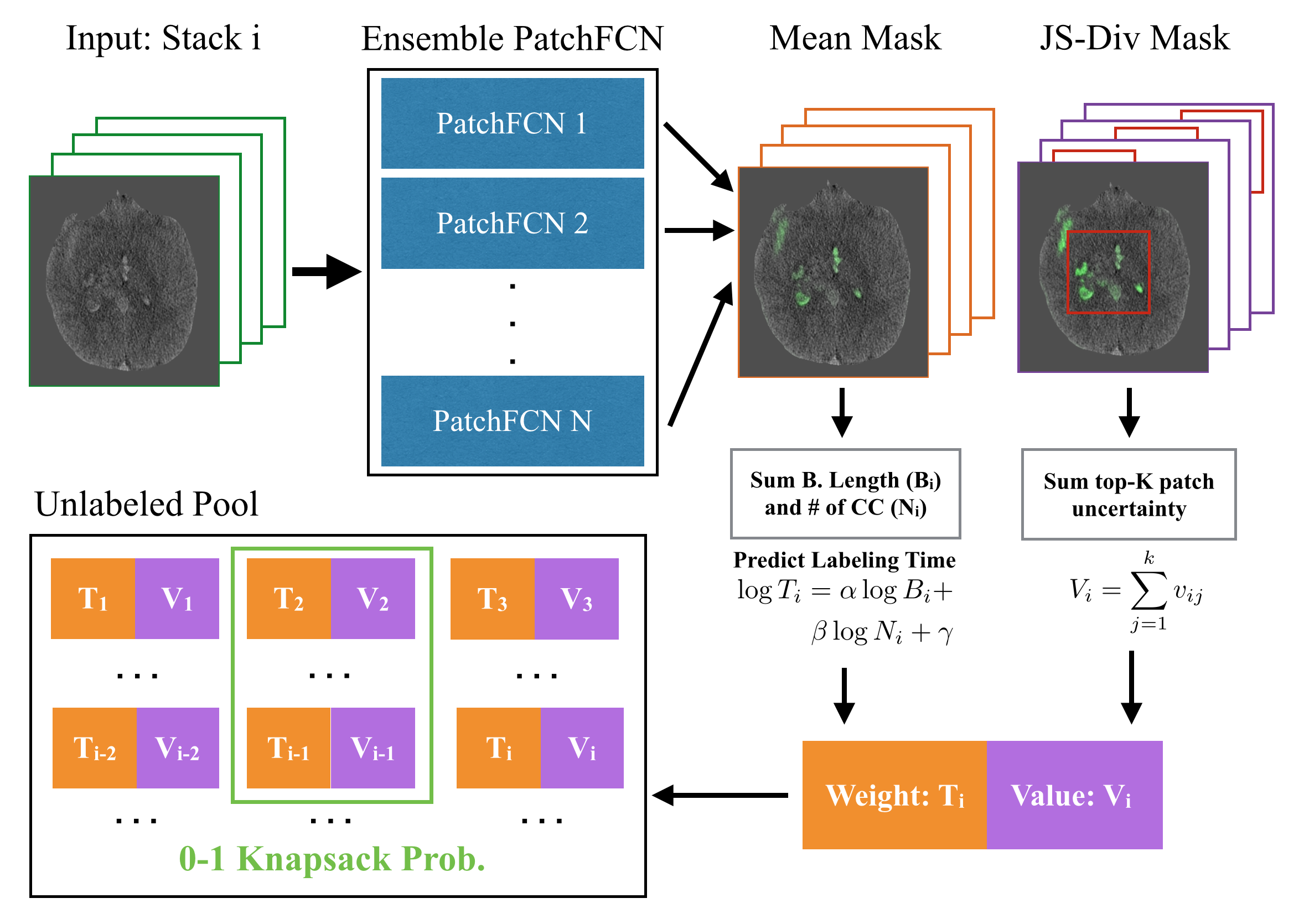}
    \vspace*{-0.3cm}
    \caption{Overview. 
    First, the stack runs through the ensemble PatchFCNs trained on the seed set $S$, which produces the mean hemorrhage heatmap and the Jensen-Shannon (JS) divergence uncertainty heatmap. From the mean hemorrhage heatmap, we apply multiple thresholds to compute the mean boundary length $B_i$ and number of connected components $N_i$. Our log-regression model then takes $B_i$ and $N_i$ to predict the stack labeling time $T_i$. The sum of uncertainty of the top-K uncertain patches is defined to be the stack uncertainty $V_i$. Given any fixed labeling budget(time) $Q$, we treat each stack in the unlabeled pool as an item of weight $T_i$ and value $V_i$. The optimal set of items for annotation is obtained by solving a 0-1 Knapsack problem with dynamic programming.
    }
    \vspace*{-0.3cm}
    \label{fig:system}
\end{figure}

\vspace*{-0.2cm}
\section{Supervised Learning System}
\vspace*{-0.2cm}

\begin{figure}[t]
    \centering
    \includegraphics[width=0.6\linewidth]{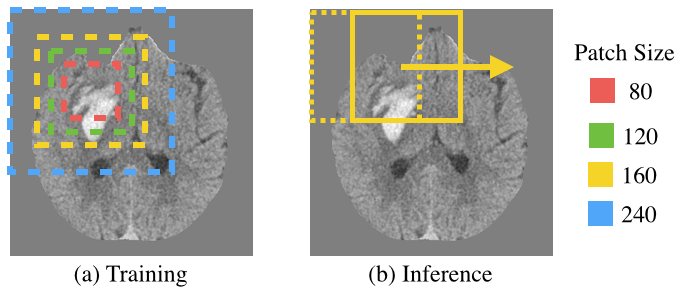}
    \vspace*{-0.3cm}
    \caption{PatchFCN system. We train the network on patches and test it in a sliding window fashion. The optimal crop size is found to be 160x160 for our task. 
    }
    \vspace*{-0.2cm}
    \label{fig:patch-fcn}
\end{figure}

As a machine learning system we use a convolutional neural network (CNN). More specifically we use a fully convolutional neural network (FCN). FCNs are able to make pixelwise predictions. The standard approach for using an FCN is to input the entire image into the FCN and obtain pixelwise predictions with a single forward pass \cite{long2015fully,yuhmukherjeemanley2015resource}. We instead use an FCN which uses a patch as input and makes predictions for presence of hemorrhage for each pixel within a specific patch at a time, which we call PatchFCN. This architecture has the advantage that the network has to make its predictions based on the local morphology and hence is less prone to overfit into the global context, which results in better test time accuracy than standard FCNs. At test time we apply the PatchFCN in a sliding window fashion (see Fig.~\ref{fig:patch-fcn}). We extensively tested this network architecture in a separate technical report \cite{arxiv2018PatchFCN} and established that it outperforms whole image baselines for various underlying FCN architectures. We use the 38 layer dilated residual net (DRN) as specific FCN architecture. It uses dilated convolutions to preserve spatial resolution together with residual connections \cite{yu2017dilated}. We also group the pixelwise predictions into regions using connected component analysis and aggregate the pixelwise predictions into frame and stack classification scores. This facilitates hemorrhage detection at the pixel, region, frame and stack level.

\vspace*{-0.2cm}
\section{Cost-sensitive Active Learning}
\vspace*{-0.2cm}

Let us define our active learning problem as follows: given a labeled seed set $S$ and an unlabeled pool set $U$, find a small subset $P$ from $U$ for labeling that maximizes a suitable test set metric. Our system which is depicted in Fig.~\ref{fig:system} estimates an uncertainty score for each example (see Sec.~\ref{sec:uncertainty}) and the labeling time (see Sec.~\ref{sect:predtime}). The goal is to select the set of examples such that the sum of their uncertainty is maximized under the constraint that the total estimated labeling time stays within a given budget. The optimal selection of items reduces to the well-known 0-1 Knapsack problem, which can be solved with dynamic programming.



\subsection{Uncertainty Measure}
\label{sec:uncertainty}
\vspace*{-0.05cm}
Uncertainty (or informativeness) is at the core of active learning techniques. It can be estimated by single model outputs~\cite{lewis1994sequential} or a committee of models~\cite{seung1992query}. The idea of query-by-committee (QBC) is to run multiple models on the same example and use their disagreement to estimate uncertainty. Experimentally, we found that QBC consistently works better than single-model uncertainty.
Within the QBC framework, we have tried various uncertainty measures and found the 
Jensen-Shannon (JS) divergence to work best. Concretely, let's assume we have $N$ models in the committee and the output distribution of model $i$ is $P_i$. The JS divergence is then defined as: 
\vspace*{-0.3cm}
\begin{equation}
  JS(P_1,P_2,...,P_N) = H(\frac{1}{N}\sum_i^{N}P_i) - \frac{1}{N}\sum_i^{N}H(P_i)
\vspace*{-0.3cm}
\end{equation}
where H is the entropy function.

We average all pixelwise uncertainties within each patch to obtain the uncertainty of a patch. The stack uncertainty is obtained by averaging the top $K$ uncertain patches within the stack. The choice of $K$ is a balance between taking the max ($K=1$) or the mean ($K=\infty$) of the whole stack. In all AL experiments in this paper, we set $K=200$ and number of models $N=4$. We have tried larger $N$ but didn't gain any performance. Visualization of such uncertainty can be found in Fig.~\ref{fig:examples-mined}.


\subsection{Labeling Time Prediction}
\vspace*{-0.05cm}
\label{sect:predtime}
First, we need to ask what is the optimal unit of labeling -- patch, frame or stack? Employing our neuro-radiology expertise, we settled on labeling stacks. While labeling patches/frames may seem more effective from a machine learning perspective, it comes with a severe overhead, i.e. the whole stacks need to be retrieved and examined by radiologists anyway. Therefore, it is less efficient than labeling the stacks.

\begin{figure}[t]
    \centering
    \includegraphics[width=0.35\linewidth]{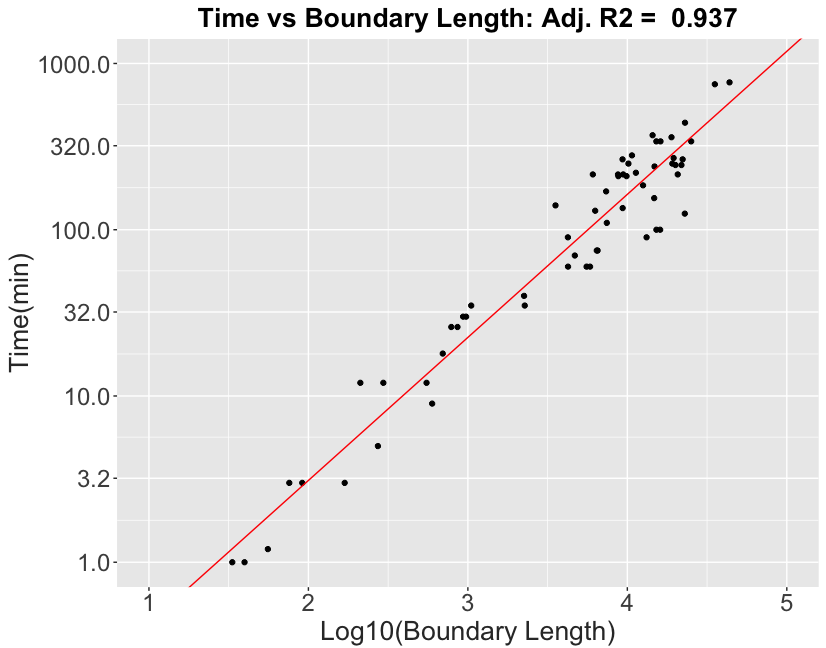}
    \includegraphics[width=0.35\linewidth]{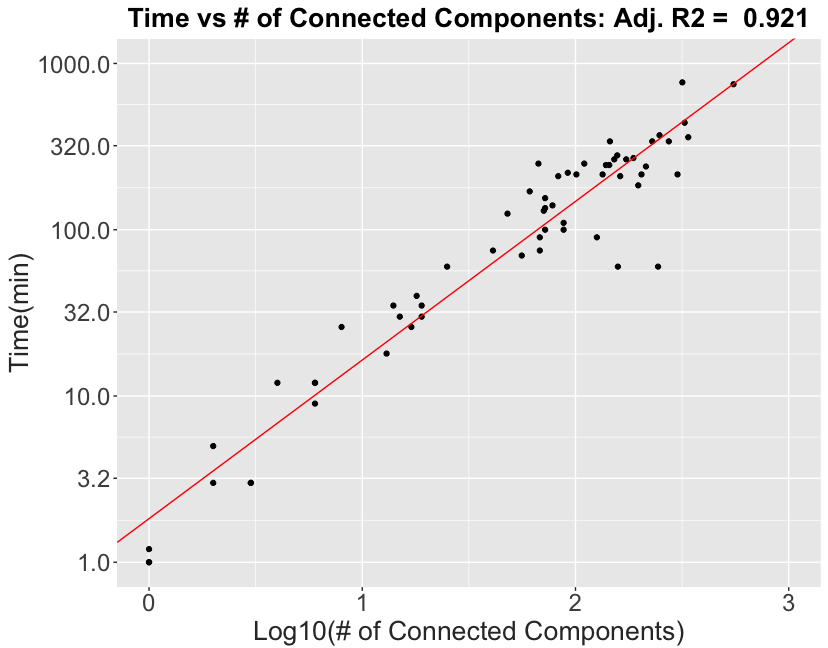}
    \caption{Left: Time vs Log(Boundary Length). Right: Time vs Log(Number of Connected Components). Both plots show the goodness of our linear fit and the normality of residuals after the log transform. Note that the y-axis is actually displayed in log-scale.}
    \label{fig:pred_label_time}
    \vspace*{-0.3cm}
\end{figure}

To apply active learning in practice, we need to ensure it actually saves labeling cost or efforts. This is crucial as per-stack labeling times in our data span $3$ orders of magnitude. We utilize linear regression to predict the log labeling time $\log{t}$ based on two features: 1) mask boundary length $B$, and 2) number of connected components $M$ under the log-transform. 
\vspace*{-0.3cm}
\begin{equation}
    \log{t} = \alpha \log{B} + \beta \log{M} + \gamma
\vspace*{-0.15cm}
\end{equation}
Fig.~\ref{fig:pred_label_time} shows the effectiveness of our log-transform and the goodness of fit on both features. 61 data points were used to fit the linear model, which we found to be sufficient. In order to compute the features at test time we use the pixelwise predictions of our network. We also tried using deep FCN features from an intermediate layer directly but found the prediction to be less stable.

\vspace*{-0.1cm}
\section{Data Collection}
\vspace*{-0.2cm}
\label{sect:data}
Our pixelwise labeled dataset contains $1247$ clinical head CT scans ($29095$ valid frames) performed from 2010-2017 on 64-detector-row CT scanners (GE, Siemens) at our affiliated hospitals. 
Each scan is a stack of $27$-$38$ frames with in-plane resolution close to $0.5$mm and z-axis resolution of $5$mm. Scans were anonymized by removing all protected health information as well as skull, scalp and face.
A board-certified neuroradiologist with specialization in traumatic brain injury (TBI) identified areas of acute intracranial hemorrhage at the pixel level.
We randomly split the dataset into a trainval/test set of $934$/$313$ stacks, called $S_{trainval}$, $S_{test}$ respectively (S for seed).

The unlabeled set was collected using key phrase searches of radiology reports. We searched independently for positive and negative cases. The search for positive cases over 1 year yielded $1755$ cases. A separate search over a shorter period identified 640 negative cases. We call this set of cases set U (for unlabeled) to be distinguished from set S. Also, $120$ randomly selected cases from U (called $U_{test}$) were annotated at stack level in order to benchmark our system in this domain.  

\begin{figure}[t]
    \centering
    \includegraphics[width=0.9\linewidth]{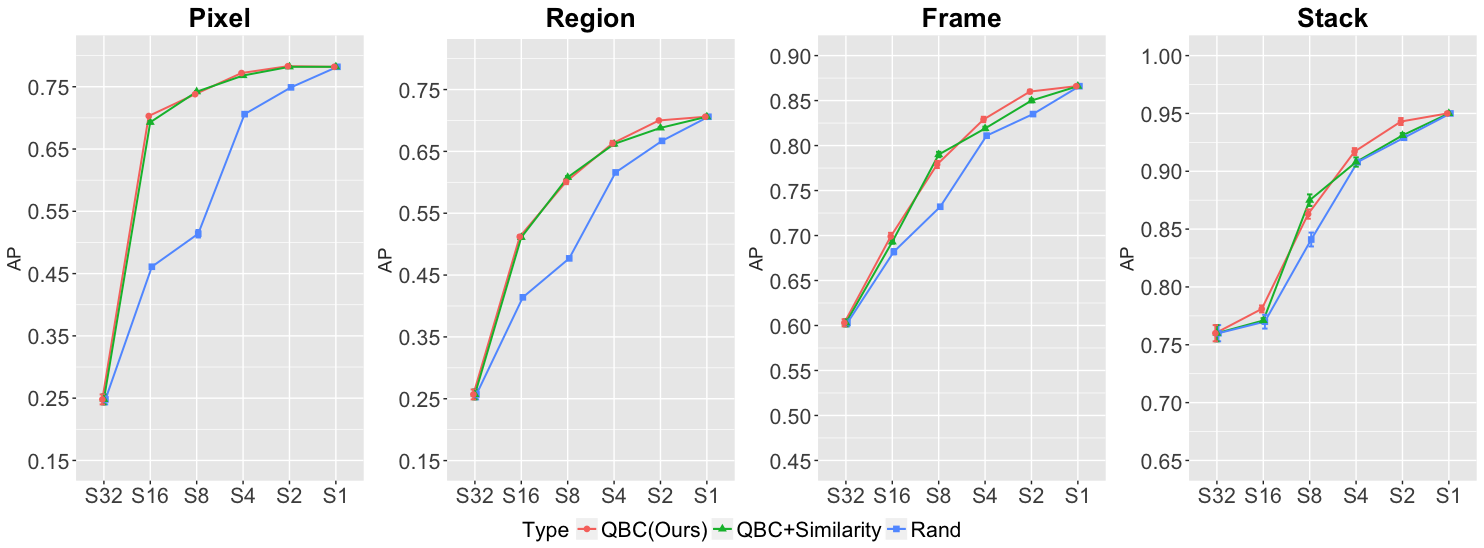}
    \caption{Core-set selection curves. Our system (QBC) starts to outperform ~\cite{yang2017suggestive} (QBC + Similarity) on region, frame and stack level as the dataset grows beyond one fourth of the whole set. Both QBC algorithms maintain a large gap with random baselines on pixel and region APs. For the frame and stack APs, our system still maintains a healthy margin above the random baseline for all data sizes. The region AP is computed following the definition of ~\cite{arxiv2018PatchFCN}.}
    \label{fig:core-set}
    \vspace*{-0.3cm}
\end{figure}

\section{Experiments}
\vspace*{-0.05cm}

\subsection{Core-set Active Learning}
\vspace*{-0.05cm}
A core-set is a subset of the training set where the empirical loss of a model is similar to that on the entire training set. In this experiment, we grow the core-set iteratively and study how the performance improves~\cite{yang2017suggestive,sener2018active}. For fair comparison, we strip away the cost prediction and Knapsack-solving part of our full system (See Fig. ~\ref{fig:system}), and select examples based on their uncertainty scores alone. 

We use the average precision (AP) metric to compare algorithms. Fig.~\ref{fig:core-set} shows the performance of our query-by-committee system (QBC), suggestive annotation system (QBC + Similarity)~\cite{yang2017suggestive}, and random baseline. In this comparison, we improve~\cite{yang2017suggestive} by using the patch-based approach for QBC + Similarity baseline, because PatchFCN~\cite{arxiv2018PatchFCN} gives better uncertainty and similarity measures than vanilla FCN. Without it, we observed a significant performance drop. Following~\cite{yang2017suggestive}, we tried diversifying the ensemble with bootstrapping, but did not see benefit. 

The experiment began with a seed set $1/32$ of the training set, and doubled it by either random sampling or active learning. In the next round, this doubled set becomes the new seed set and the process repeats. In each round, we trained an ensemble for all methods in order to compute QBC uncertainty. Fig.~\ref{fig:core-set} shows that our system's performance at half the dataset (S2) closely matches the performance of using the whole dataset (S1) for every AP, similar to ~\cite{yang2017suggestive,sener2018active}. However, here we use a dataset that is two orders of magnitude larger and much harder to overfit on. 

Our experiment indicates that on a large dataset, QBC uncertainty alone could be sufficient to yield competitive performance, if not state-of-the-art. Without bootstrapping or pairwise similarity, our system beats the random baseline by a good margin and compares favorably with ~\cite{yang2017suggestive} in performance and time complexity. The time complexity of core-set approaches~\cite{yang2017suggestive,sener2018active} are dominated by the pairwise similarity computation, which is quadratic and can be expensive in practice when the seed set is too large to be grown by brute-force labeling. In contrast, our system has linear time complexity because it computes everything on-the-fly.

\subsection{Cost-Sensitive Active Learning}
\vspace*{-0.05cm}

\begin{figure}[t]
    \centering
    \includegraphics[width=0.9\linewidth]{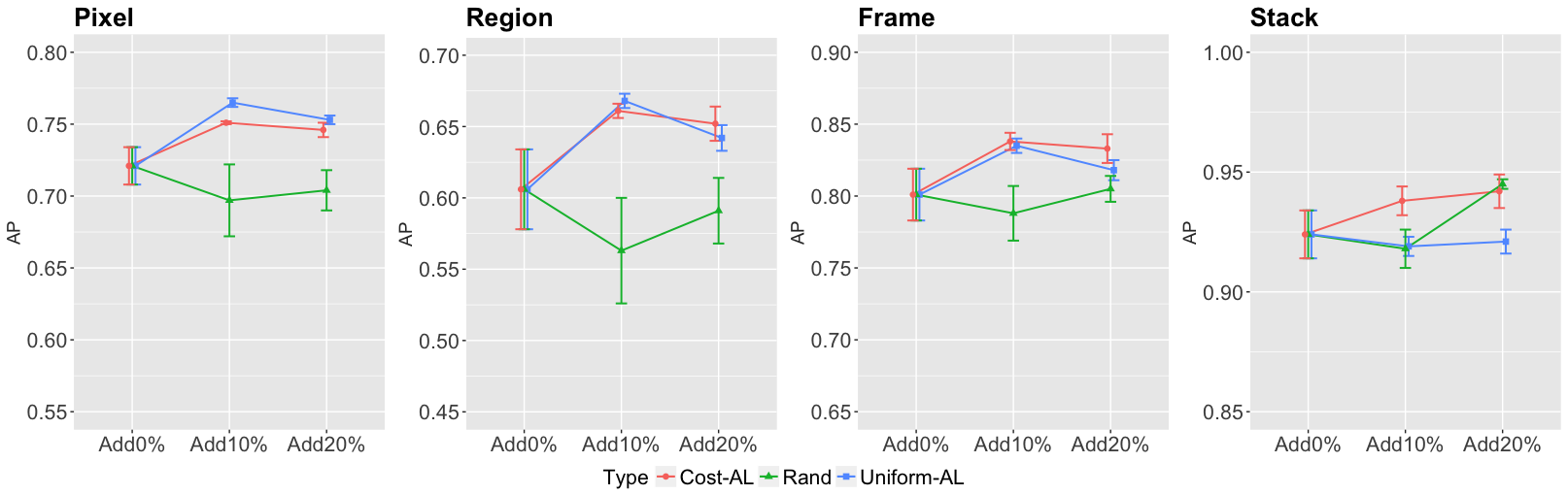}
    \vspace*{-0.3cm}
    \caption{Cost-sensitive active learning. At the first iteration, the system achieves much better performance than the random baseline for all metrics. The random baseline does not improve over the seed set. In the next round, the random baseline improves the stack AP while the ALs remain the same. The error bars of AL come from the network initialization and the stochastic gradient (SGD) training. The error bars of random baseline mostly come from the random addition of data, plus the same sources of AL randomness. The time increment is $10\%$ of the total labeling time of the pool, which simulates the situation where our budget is only a small fraction of the total labeling cost.}
    \label{fig:cost-sensitive}
    \vspace*{-0.05cm}
\end{figure}

After validating the core-set AL, we model the cost with the full system described in Fig.~\ref{fig:system}. We randomly select half of our labeled training set as the seed set to mimic the scenario where the seed set is large enough to render naive labeling impractical for growing the data. Yet at the same time we want the pool to be at least as large as the seed. In each iteration, we increment the data by allocating additional \textit{time} to add labeled examples by solving the Knapsack problem. For the random baseline, we randomly select examples to add until no example can fit in the given time anymore. Fig.~\ref{fig:cost-sensitive} shows the superiority of our system (QBC) over both uniform-cost AL (UAL) and the random baseline in such setting. The result supports Fig.~\ref{fig:examples-mined} where UAL is biased toward examples with large bleeds and long labeling times. In fact, UAL selected 8/11 stacks in the first/second rounds, whereas cost-sensitive AL (CAL) selected 94/107 stacks. Due to lack of stack diversity, UAL performs worse than CAL at the stack level. 

The strong gain of CAL at (+10\%) not carrying over to (+20\%) is explained by the ratio of unlabeled pool to the labeled training set. When the ratio is small, the data is insufficient for AL system to choose from. In Fig.~\ref{fig:core-set}, the ratio starts with 3100\% and stops with 100\% at S2. In Fig.~\ref{fig:cost-sensitive}, the ratio started with 100\%. After (+10\%) round, the ratio is 66\% for CAL and 80\% for Rand. The leveling off of CAL performance shows that most of the informative examples were already selected in the (+10\%) round.

\subsection{Active Learning in the Wild}
\vspace*{-0.05cm}

\begin{figure}[t]
    \centering
    \includegraphics[width=0.9\linewidth]{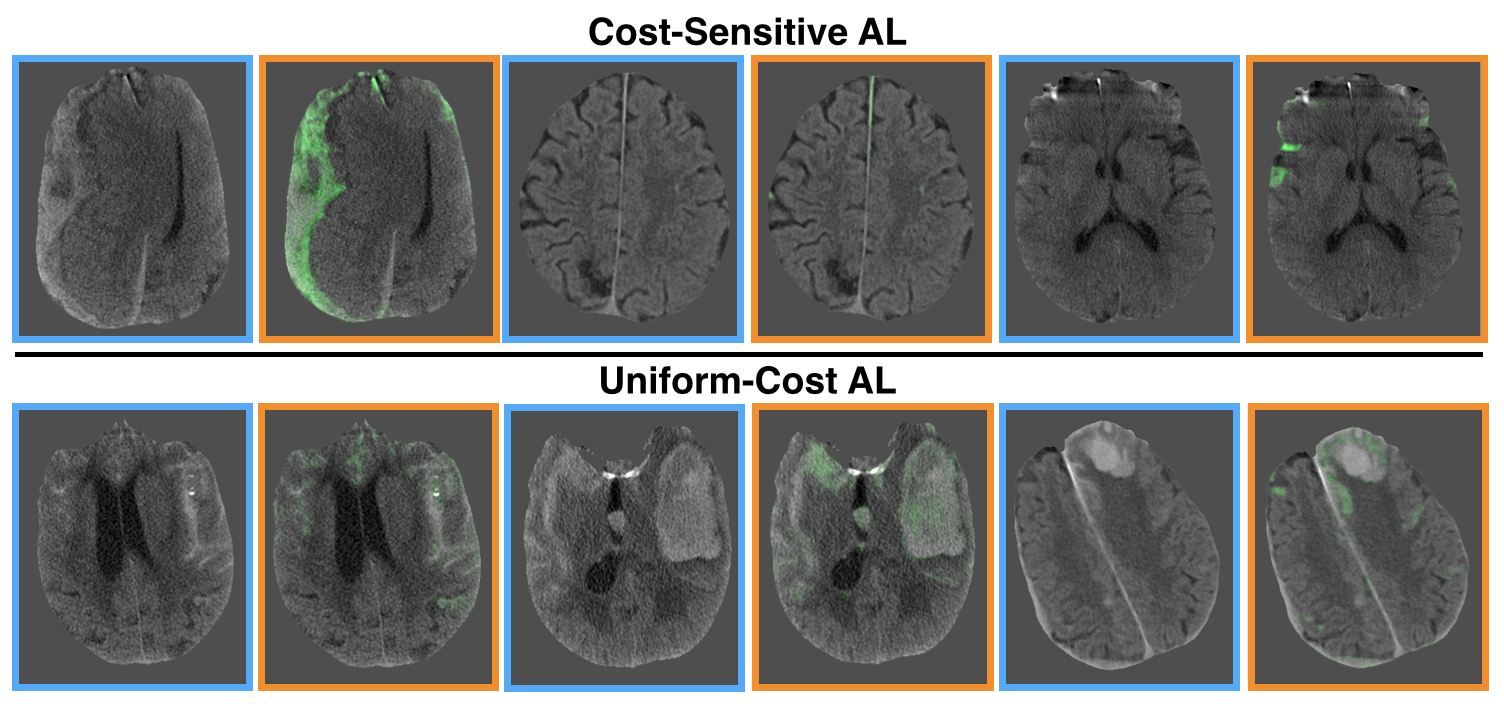}
    \caption{Examples selected by cost-sensitive and uniform-cost AL systems. Blue boxes are the original images, while orange boxes are the images overlaid with Jensen-Shannon divergence. The brightness of the green color indicates uncertainty. The examples selected by uniform-cost system mostly contain massive bleeds and are substantially more time-consuming for annotation, whereas examples by the cost-sensitive system are diverse and meaningful, maximizing the return on investment.}
    \label{fig:examples-mined}
    \vspace*{-0.5cm}
\end{figure}

Finally, we apply our system on the unlabeled pool described in Sec.~\ref{sect:data}. First, we train an ensemble on the entire labeled set. Then we select examples from the unlabeled pool under a budget of $100$ hours. A neuroradiologist examined the selected cases and determined there were $115$ negatives and $64$ positives. There were also $51$ subacute or postsurgical cases we excluded. The actual labeling time turned out to be within $10\%$ of our estimate. We call these newly annotated examples $U_{train}$, to be distinguished from $S_{trainval}$ defined in Sec.~\ref{sect:data}. To qualitatively assess the impact of cost modeling, we show examples mined by both uniform-cost and cost-sensitive AL in Fig.~\ref{fig:examples-mined}.

For quantitative benchmarking, we trained an ensemble of 4 PatchFCNs from scratch with the newly augmented data (Ensemble $S_{trainval}$+$U_{train}$) and compared them with the ensemble trained on the original data (Ensemble $S_{trainval}$). The results on $S_{test}$ and $U_{test}$ are shown in Table.~\ref{table:unlabel}. We benchmark on two test sets here because we care about the performance on both seed $S$ and pool $U$ domains, which in practice are often not exactly the same. The gain on $S_{test}$ shows that our method works despite the domain shift, and the strong gain on $U_{test}$ demonstrates how a model trained on large data can be improved by collecting a little more data judiciously. 
\begin{table}[t]
     \centering
     \def\arraystretch{1.25}\tabcolsep=3pt
     \begin{tabular}{|c | c c |} 
     \hline
     $S_{test}$ & Pixel AP & Stack AP\\
     \hline\hline
     Ens. $(S \cup U)_{train}$	& $77.9 \pm 0.3\%$ & $\mathbf{95.6 \pm 0.9\%}$\\
     \hline
     Ens. $S_{trainval}$ & $\mathbf{78.2 \pm 0.2\%}$ & $95.0 \pm 0.1\%$ \\
     \hline
     \end{tabular}
     \begin{tabular}{|c | c |} 
     \hline
     $U_{test}$ & Stack AP\\
     \hline\hline
     Ens. $(S \cup U)_{train}$	& $\mathbf{90.1 \pm 1.7\%}$ \\
     \hline
     Ens. $S_{trainval}$ & $85.1 \pm 0.3\%$ \\
     \hline
     \end{tabular}         
    \caption{Left: Performance on $S_{test}$. Compared to Ensemble $S_{trainval}$, Ensemble $(S \cup U)_{train}$ performs just as well on the pixel level and slightly outperform on the stack level. Right: Performance on $U_{test}$. Ensemble $(S \cup U)_{train}$ beats Ensemble $S_{trainval}$ by a good margin on the pool set.}
    \label{table:unlabel}
    \vspace*{-0.8cm}
\end{table}



\vspace*{-0.15cm}
\section{Conclusion}
\vspace*{-0.25cm}
In this paper, we proposed a cost-sensitive, query-by-committee active learning system for intracranial hemorrhage detection. We validated it on a substantially larger pixelwise labeled dataset than earlier works and applied it to improve the model by annotating new data from the wild. Our study demonstrates the potential of growing large medical  datasets to the next level with cost-sensitive active learning. 

\vspace{0.1cm}

\noindent \textbf{Acknowledgments:} This work was supported in part by California Initiative to Advance Precision Medicine. Christian H{\"a}ne received funding from the Swiss National Science foundation (165245). Amazon Web Services provided part of the compute time.

{\small
\bibliographystyle{splncs}
\bibliography{egbib}
}

\end{document}